\documentclass[letterpaper]{article}
\usepackage[preprint]{aaai2027}
\usepackage[hyphens]{url}
\usepackage{graphicx}
\usepackage{natbib}
\usepackage{caption}
\usepackage{algorithm}
\usepackage{algorithmic}
\usepackage{booktabs}
\usepackage{amsmath, amssymb, amsthm}

\newtheorem{definition}{Definition}
\newtheorem{proposition}{Proposition}
\newtheorem{lemma}{Lemma}
\newtheorem{corollary}{Corollary}
\newtheorem{theorem}{Theorem}

\title{Kant-Q: A Q-Learning Algorithm for Kantian Equilibrium\\ in Symmetric Markov Games}
\author{
    Rishi Bharadwaj\textsuperscript{\rm 1}\corresponding,
    Sujit Gujar\textsuperscript{\rm 2},
    Yadati Narahari\textsuperscript{\rm 1}
}
\affiliations{
    \textsuperscript{\rm 1}Department of Computer Science and Automation, Indian Institute of Science, Bengaluru\\
    \textsuperscript{\rm 2}International Institute of Information Technology, Hyderabad\\
    rishib62107@gmail.com
}

\begin{document}

\maketitle

\begin{abstract}
AI agents increasingly interact with one another, and many of these interactions are mixed-motive games, which have posed a persistent challenge for learning algorithms. Most multi-agent reinforcement learning (MARL) algorithms are designed to reach Nash equilibria, yet in many such games these equilibria are Pareto-inefficient. We instead study Kantian equilibria, an economic solution concept in which each agent plays the action it would want all agents to adopt. In symmetric games, this corresponds to an optimal common policy. We propose Kant-Q, a Q-learning algorithm in which each agent learns action profiles where all agents play the same action. Each update is then a sample from a single-agent MDP, so Kant-Q inherits the convergence guarantees of standard Q-learning. We show that every symmetric Markov game has a pure Kantian equilibrium. We also prove that all agents' Q-tables converge to the same limit, whose greedy policy is such an equilibrium, and that greedy play converges whenever the optimal common action is unique at every state. We further show that in a subclass of symmetric monotone games, the Kantian equilibrium is Pareto-efficient among pure strategy profiles and strictly individually rational. Empirically, on the iterated Prisoner's Dilemma and the GovSim fishery with up to 200 agents, Kant-Q learns the Kantian equilibrium, while the baselines underperform. Since copies of the same AI system naturally face one another in symmetric games, these results suggest universalization as a design principle for cooperative AI--AI interaction.
\end{abstract}

\section{Introduction}
Reaching Pareto-efficient outcomes in mixed motive games has been a longstanding challenge in the world of Multi Agent Reinforcement Learning (MARL). Most MARL algorithms focus on learning Nash-equilibria~\citep{nashq,conitzer2007}, which in many general-sum games, are Pareto-inefficient~\citep{conitzer_foundations}. Other approaches, such as Multi-Agent PPO~\citep{MAPPO} can lead to unfair outcomes, where a subset of agents receive low payoffs to maximise the overall reward. The problem is increasingly practical, as AI systems are deployed as autonomous agents that interact with one another, for example when they negotiate, trade, or draw on a shared resource~\citep{conitzer_foundations, govsim}. As an alternate solution concept, the economist~\citet{roemer} proposes Kantian equilibrium, based on the paradigm, “What is the strategy I would like all of us to play?” There is evidence that humans actually reason this way in social dilemma games~\citep{Levine}. Interestingly, a similar result has also been found in LLMs, where ``Universalization''-based reasoning allowed LLM agents to achieve significantly better performance in GOVSIM~\citep{govsim}.

This kind of decision making rule is not new. Superrationality, put forward by~\citet{Hofstadter1983}, follows a near-identical paradigm: “If reasoning guides me to say C, then, as I am no different from anyone else as far as rational thinking is concerned, it will guide everyone to say C.” Functional Decision Theory~\citep{yudkowsky2018fdt} follows similar reasoning when facing the twin Prisoner's Dilemma: “My twin and I follow the same course of reasoning—this one. The question is how this very course of reasoning should conclude. If it concludes that cooperation is better, then
we both cooperate and I get \$1,000,000. If it concludes that defection is better,
then we both defect and I get \$1,000. I would be personally better off in the former
case, so this course of reasoning hereby concludes cooperate.” For AI agents this premise can hold literally, as interacting agents can share the same source code. By contrast, standard Q-learning treats other agents as a stationary part of the environment~\citep{hernandez}, so Q-learners behave more like agents following Causal Decision Theory~\citep{bell2021reinforcement}.

Kantian reasoning as an economic concept offers a fair, efficient way of resolving conflicting strategies between symmetric players. Although it was introduced to explain cooperation in humans, it is equally applicable to AI--AI interaction. For example, AI agents could act as proxies for large numbers of humans in a game whose payoffs are not known in advance, and learn the Kantian equilibrium on their behalf (Section~\ref{sec:limitations}). As our experiments show, with shared exploration it is a scalable method of deploying large numbers of agents that then reach a cooperative equilibrium, without needing prior centralised training. Every agent receives the same payoff, preventing an unfair distribution of rewards. This is in stark contrast to many cooperative algorithms that maximise total welfare. While it is vulnerable to defection, it nevertheless offers an interesting approach to large-scale agentic cooperation.

To the best of our knowledge, there are no learning algorithms built explicitly around the Kantian or superrational counterfactual. To address this gap, we introduce Kant-Q, a Q-learning algorithm built around the Kantian counterfactual, particularly for symmetric Markov games. The agents do not know the game, and observe only the state, their own reward and a single match bit indicating whether all agents played the same action. If the game were known, the SKE could be computed directly by solving a single-agent MDP, as our value-iteration baseline does (Section~\ref{sec:experiments}). The agents' Q-tables are updated only when the match bit is true. The simple but powerful observation is that every such update is equivalent to a sample from a single-agent MDP whose actions are the diagonal profiles, the joint actions in which every agent plays the same action. Kant-Q therefore does not suffer from the non-stationarity of independent Q-learning, since the other agents affect only which entries are updated and how often, not the expected update target. Convergence then follows from standard results on Q-learning~\citep{tsitsiklis1994}. Our contributions are as follows.
\begin{itemize}
    \item \textbf{Existence of equilibrium.} We first show that the Kant-Q operator is a contraction (Lemma~\ref{lem:contraction}). Together with the optimality of its fixed point, this shows that every symmetric Markov game has a pure simple Kantian equilibrium (SKE). A deterministic policy is an SKE if and only if it is optimal in the single-agent MDP above (Theorem~\ref{thm:kantq}).
    \item \textbf{A convergent learning algorithm.} In symmetric Markov games, Lemma~\ref{lem:contraction} and the convergence theorem of~\citet{tsitsiklis1994} show that every agent's Kant-Q table converges with probability $1$ to the same limit, whose greedy policy is a pure SKE (Theorems~\ref{thm:convergence} and~\ref{thm:kantq}). When the optimal action is unique at every state, the agents' greedy actions settle on it in finite time (Corollary~\ref{cor:play}). With shared exploration, the sample complexity of learning the SKE is only polylogarithmic in the number of agents (Proposition~\ref{prop:rate}).
    \item \textbf{Fairness guarantees.} In a subclass of~\citet{roemer}'s strictly monotone games, the SKE is Pareto-efficient among pure strategy profiles and strictly individually rational (Propositions~\ref{prop:ske-exists} and~\ref{prop:ir}). In symmetric games, it is also fair, as all agents receive the same payoff.
    \item \textbf{Experiments.} On the iterated Prisoner's Dilemma and the GovSim fishery with up to 200 agents, Kant-Q learns the SKE, cooperating where independent Q-learning and Friend-Q defect, and sustaining the fishery with equal catches, while the baselines underperform.
\end{itemize}

\section{Related Work}
\paragraph{Cooperative MARL.}
Cooperation in MARL is a very active field. For an introduction, see~\citet{du2023reviewcooperationmultiagentlearning}. Building on their taxonomy, we group cooperative approaches into six categories. \emph{Other-regarding preferences} reshape agents' rewards to value others' welfare~\citep{prosocial, inequity_aversion, altruistic_behaviour}. \emph{Opponent-shaping} methods account for how an agent's actions influence co-players' learning~\citep{foerster_lola, letcher2018stable,fcl}. Among these, Foolproof Cooperative Learning (FCL)~\citep{fcl} is closest in goal, reaching individually rational, efficient outcomes in symmetric stochastic games by retaliating against deviators. However, it requires joint-action value functions, a communicated exploration seed and repeated interaction. Kant-Q needs no joint-action values, cooperates even in one-shot games, and converges without a shared seed, though it uses one to scale to many players (Section~\ref{sec:shared}), at the cost of FCL's robustness to deviating agents. \emph{Reputation and norms} methods track and sanction co-players' behaviour~\citep{mckee2023multiagentreinforcementlearningmodel, coop_nicolas}. \emph{Contract} methods extend the game with binding commitments or reward transfers~\citep{hughes_learning, get_it_in_writing}. \emph{Satisficing} learners retain actions that meet an adaptive aspiration level~\citep{stimpson_2003, Chasparis_2013, marden}. Finally, \emph{conditional learners} condition their beliefs about co-players on their own actions or on observed similarity~\citep{banerjee2007, Kuroe2024ClassesOD, oesterheld_similarity}. Kant-Q is closest to this last category. Unlike these methods, however, it does not learn the correlation between agents empirically. Instead, it builds universalization directly into the value update. Moreover, satisficing and conditional learners typically steer agents toward the welfare-maximizing profile, which can sacrifice individual agents for the collective. They are also generally only tested on single state games. In contrast, the Kantian equilibrium of a symmetric game gives all agents equal rewards, and Kant-Q applies to symmetric Markov games.

\paragraph{Q-learning in Markov games.}
Kant-Q also belongs to the family of Nash-Q~\citep{nashq} and Friend-Q~\citep{littman2001friend}, which extend Q-learning to Markov games by replacing the max in the Q-learning target. Nash-Q uses the value of a Nash equilibrium of this stage game, while Friend-Q uses the maximum over all joint actions, under the assumption that all other agents assist. In the Prisoner's Dilemma, both lead to mutual defection. The only Nash equilibrium is for both agents to defect, and each Friend-Q agent's best joint action is to defect while the other cooperates, so both defect. A small change to Friend-Q yields cooperation. Kant-Q takes the same maximum, but only over diagonal profiles. This removes the profiles in which the agent exploits the others, and the best remaining profile is mutual cooperation.

\section{Setting and Algorithm}
\label{sec:setting}

We first discuss properties of single-state games, before extending the discussion to Markov games in Section~\ref{sec:markov}.

\subsection{Symmetric games}
\label{sec:symgames}

Consider an $N$-player game, $N \geq 2$, with a common, finite, ordered action set $A = \{a_1, a_2, \ldots, a_m\}$, $m \geq 2$. The game $\mathcal{G}$ is defined by payoff functions $r^1, \ldots, r^N : A^N \to \mathbb{R}$, where $r^i(b_1, \ldots, b_N)$ denotes player $i$'s payoff when each player $\ell$ plays $b_\ell$. We write $b_{-i} \in A^{N-1}$ for the profile of the $N-1$ players other than $i$, and $r^i(b_i \, ; b_{-i})$ for player $i$'s payoff when they play $b_i$ and their opponents play $b_{-i}$. Throughout, $r^i(b \, ; a, \ldots, a)$ denotes the payoff to a player who plays $b$ while all $N-1$ opponents play $a$, and $r(\bar a)$ the payoff at the \emph{diagonal profile} $\bar a := (a, \ldots, a) \in A^N$, in which all $N$ players play $a$. Given an opponent profile $c \in A^{N-1}$, a coordinate $k$, and an action $a \in A$, we write $(c_{-k}, a)$ for the opponent profile obtained from $c$ by replacing its $k$-th coordinate with $a$ and leaving the rest unchanged.

We call a game \emph{symmetric} if it is symmetric with respect to every permutation of the players, i.e., a totally symmetric game~\citep{vonneumann}. That is, for every bijection $\sigma : \{1, \dots, N\} \to \{1, \dots, N\}$, every profile $\mathbf{b} = (b_1, \dots, b_N) \in A^N$, and every player $i$,
\[
  r^{\sigma(i)}(b_1, \dots, b_N) = r^i\bigl(b_{\sigma(1)}, \dots, b_{\sigma(N)}\bigr).
\]
In words, permuting the players' actions permutes their payoffs in the same way, so a player's payoff does not depend on who they are or on which opponent plays which action. Under symmetry we therefore drop the player index and write
\[
    r(b \, ; c) := r^i(b \, ; c), \qquad b \in A, \; c \in A^{N-1},
\]
which does not depend on $i$. For $N = 2$, symmetry reduces to the usual condition that for all $a_j, a_l \in A$,
\[
    r^1(a_j, a_l) = r^2(a_l, a_j).
\]

\begin{definition}[Strictly Monotone Symmetric Game]
\label{def:monotone}
A symmetric game $\mathcal{G}$ is \emph{strictly monotone decreasing} if:
\begin{enumerate}
    \item \textbf{(Monotonicity)} $r$ is strictly decreasing in each opponent's action, holding the remaining opponents fixed. That is, for every $b \in A$, every opponent profile $c \in A^{N-1}$, every coordinate $k$, and all $j < l$,
    \[
        r(b \, ; (c_{-k}, a_j)) > r(b \, ; (c_{-k}, a_l)).
    \]

    \item \textbf{(Common Diagonal Ordering)} The diagonal payoffs are strictly decreasing in the common action. That is, for all $j < l$,
    \[
        r(\bar a_j) > r(\bar a_l).
    \]
\end{enumerate}
\end{definition}

\noindent Clause~1 is the strict monotonicity condition of~\citet{roemer}. Clause~2 is an additional restriction we impose, requiring the diagonal payoffs to be strictly ordered in the same direction as monotonicity. Reversing the order of the action indices gives a \emph{strictly monotone increasing} game. We call a game \emph{strictly monotone} if it is either, and throughout the paper this term refers to this class, with Clause~2 included. The Prisoner's Dilemma is strictly monotone decreasing, with $a_1 = C$ (cooperate) and $a_2 = D$ (defect).

\begin{definition}[Simple Kantian Equilibrium]
\label{def:ske}
In a symmetric game, the action that each player would prefer \emph{all} to play is a \emph{simple Kantian equilibrium} (SKE)~\citep{roemer}.

Formally, in an $N$-player symmetric game with common action set $A$, an action $a^* \in A$ is an SKE if for all $a \in A$, $a \neq a^*$,
\[
    r(\bar a^*) \geq r(\bar a).
\]
\end{definition}

\noindent Since $A$ is finite, an SKE always exists. It is any maximiser of $r(\bar a)$ over $a \in A$.

\begin{proposition}[Pareto-Efficiency of an SKE~\citep{roemer}]
\label{prop:ske-exists}
In a game satisfying~\citet{roemer}'s strict monotonicity condition (Clause~1 of Definition~\ref{def:monotone}, or its increasing counterpart), any SKE is Pareto efficient.
\end{proposition}

In particular, Proposition~\ref{prop:ske-exists} applies to every strictly monotone game. By Clause~2 the SKE is unique. It is $a_1$ in the monotone decreasing case and $a_m$ in the monotone increasing case. This Pareto efficiency holds only with respect to other pure strategies, and the SKE can be Pareto dominated by mixed strategy profiles.

\begin{proposition}
\label{prop:ir}
In a strictly monotone game, the payoff profile of the unique SKE is strictly individually rational. Each player's payoff at the SKE strictly exceeds their pure-strategy minmax value.
\end{proposition}

\noindent The proof is in Appendix~\ref{app:proofs}.

\subsection{Symmetric Markov games}
\label{sec:markov}

Section~\ref{sec:symgames} considered a single stage game. Many interactions between AI agents unfold over time, and the payoff of an action depends on a state that earlier actions have shaped. We therefore extend the model to Markov games, in which the stage game depends on a state. A Markov game is a tuple $(N, S, A, P, (r^i)_{i=1}^N, \gamma)$, where $N$ is the number of players, $S$ is a finite set of states, $A$ is the common action set, $P(s' \mid s, \mathbf{b})$ is the probability of moving to state $s'$ when the joint action $\mathbf{b} \in A^N$ is played at state $s$, $r^1, \ldots, r^N : S \times A^N \to \mathbb{R}$ are the expected rewards, and $\gamma \in [0, 1)$ is the discount factor. Given the state $s$, the joint action $\mathbf{b}$ and the history, the next state is drawn from $P(\cdot \mid s, \mathbf{b})$, and player $i$'s realised reward, written $r^i_t$ at step $t$, is bounded in absolute value by a constant $R_{\max}$ and has expectation $r^i(s, \mathbf{b})$. The history includes the agents' internal randomness, such as their initial tables and exploration draws. We call the Markov game \emph{symmetric} if its stage game is symmetric in the sense of Section~\ref{sec:symgames} at every state. That is, for every permutation $\sigma$ of the players, every state $s$ and every profile $\mathbf{b} = (b_1, \ldots, b_N)$,
\[
  r^{\sigma(i)}(s, b_1, \ldots, b_N) = r^i\bigl(s, b_{\sigma(1)}, \ldots, b_{\sigma(N)}\bigr).
\]
As in the stage game, symmetry implies that all players receive the same expected reward at a diagonal profile, and we write this common value as $r(s, \bar a)$.

\paragraph{Simple Kantian equilibrium in Markov games.}

\begin{definition}[Pure SKE of a Markov game]
\label{def:markov-ske}
For a deterministic policy $\pi : S \to A$, let $V^i_{\pi}(s)$ denote player $i$'s expected discounted return from state $s$ \emph{when every player follows $\pi$}. A deterministic policy $\pi^\ast$ is a \emph{pure simple Kantian equilibrium} if, for every player $i$, every state $s$ and every deterministic policy $\pi$,
\[
  V^i_{\pi^\ast}(s) \geq V^i_{\pi}(s).
\]
\end{definition}

\noindent That is, among all deterministic policies that every player could follow simultaneously, no player prefers another one to $\pi^\ast$, from any starting state.

\subsection{Kant-Q algorithm}
\label{sec:kantq}

Each agent $i$ holds a table $Q^i_t$ with one entry $Q^i_t(s, \bar a)$ per state $s$ and diagonal profile $\bar a$, so the table has $|S| \cdot m$ entries (Algorithm~\ref{alg:kantq}). After each step, the agent observes a match bit $\mu_t$, which equals $1$ if every agent played the same action as itself and $0$ otherwise. On a matched step the joint action is the diagonal profile $\bar a^i_t$, so the agent updates only that entry. Otherwise it leaves its table unchanged. This can be seen as setting the step size $\alpha_t(s, \bar a)$ to $0$ for all unmatched steps. Each agent never observes the other agents' rewards, their individual actions or their tables, nor the reward functions and transition probabilities of the game. The tables are initialised at random and independently, so the learners are not exact copies of one another. Since Kant-Q updates only on diagonal profiles, it avoids the non-stationarity that affects independent Q-learning, as the other agents affect only when it updates.

\begin{algorithm}[t]
\caption{Kant-Q for agent $i$}
\label{alg:kantq}
\begin{algorithmic}
  \STATE {\bfseries Input:} exploration rate $\varepsilon \in (0, 1]$, learning-rate exponent $\omega \in (\tfrac12, 1)$, discount factor $\gamma$
  \STATE {\bfseries Initialise:}
  \STATE \quad Let $t = 0$ and observe the initial state $s_0$.
  \STATE \quad For all $s \in S$ and $a \in A$, draw $Q^i_0(s, \bar a)$ at random
  \STATE \quad\quad from a bounded interval, independently of the other agents,
  \STATE \quad\quad and set the update count $n(s, \bar a) := 0$.
  \STATE {\bfseries Loop:}
  \STATE \quad \textit{Choose action ($\varepsilon$-greedy):}
  \STATE \quad\quad with probability $\varepsilon$, draw $a^i_t$ uniformly at random from $A$
  \STATE \quad\quad otherwise, $a^i_t = \arg\max_{a \in A} Q^i_t(s_t, \bar a)$, breaking ties by the lowest index
  \STATE \quad \textit{Observe:} own reward $r^i_t$, next state $s_{t+1}$, and the match bit
  \STATE \quad\quad $\mu_t = 1$ if every agent played $a^i_t$, and $\mu_t = 0$ otherwise
  \STATE \quad \textit{Step size (from the match bit),} with $a := a^i_t$:
  \IF{$\mu_t = 1$}
    \STATE \quad $n(s_t, \bar a) := n(s_t, \bar a) + 1$
    \STATE \quad $\alpha_t = n(s_t, \bar a)^{-\omega}$
  \ELSE
    \STATE \quad $\alpha_t = 0$
  \ENDIF
  \STATE \quad \textit{Update:}
  \STATE \quad\quad $Q^i_{t+1}(s_t, \bar a) = (1 - \alpha_t)\, Q^i_t(s_t, \bar a)$
  \STATE \quad\quad\quad $+\, \alpha_t \bigl[\, r^i_t + \gamma \max_{a' \in A} Q^i_t(s_{t+1}, \bar a') \,\bigr]$
  \STATE \quad Let $t := t + 1$.
\end{algorithmic}
\end{algorithm}

\section{Convergence of Kant-Q}

The analysis rests on a simple but powerful observation. Whenever an agent updates its table, it receives a sample from a single-agent MDP $M_K$ whose actions are the diagonal profiles. Each agent therefore runs asynchronous Q-learning on $M_K$, and convergence and optimality follow from standard results on Q-learning~\citep{tsitsiklis1994} and MDPs~\citep{puterman1994markov}. We then show that this optimal strategy is also the simple Kantian equilibrium.

\paragraph{The diagonal MDP.}
Let $M_K$ be the single-agent MDP with states $S$, actions $A$, transitions $P_K(s' \mid s, a) := P(s' \mid s, \bar a)$, rewards $r_K(s, a) := r(s, \bar a)$ and discount factor $\gamma$, in which choosing action $a$ stands for every player choosing $a$. Since all agents observe the same state and $r(s, \bar a)$ is common to all players, $M_K$ is the same for every agent. On a matched step every agent played the same action $a$, so the next state is drawn from $P_K(\cdot \mid s_t, a)$ and each agent's reward has expectation $r_K(s_t, a)$. The step is thus a sample from $M_K$. Each agent therefore runs asynchronous Q-learning on $M_K$, with step size zero on unmatched steps. Let $\mathcal{Q}$ be the space of tables $Q$ with one real entry $Q(s, \bar a)$ per state $s$ and action $a$, with the sup norm $\| Q \| := \max_{s, a} |Q(s, \bar a)|$. The \emph{Kant-Q operator} (Algorithm~\ref{alg:kantq})  $H : \mathcal{Q} \to \mathcal{Q}$ is
\begin{equation}
\label{eq:kantq-op}
  (HQ)(s, \bar a) := r(s, \bar a) + \gamma \sum_{s' \in S} P(s' \mid s, \bar a) \, \max_{a' \in A} Q(s', \bar a').
\end{equation}
This is also the Bellman optimality operator of $M_K$~\citep{puterman1994markov}. We state its contraction property formally.

\begin{lemma}[Contraction]
\label{lem:contraction}
The Kant-Q operator $H$ is a $\gamma$-contraction in the sup norm, that is, $\| HQ - HQ' \| \leq \gamma \| Q - Q' \|$ for all $Q, Q' \in \mathcal{Q}$. Hence $H$ has a unique fixed point $Q^\ast \in \mathcal{Q}$.
\end{lemma}

\noindent The proof is in Appendix~\ref{app:proofs}.

\begin{theorem}[Convergence of Kant-Q]
\label{thm:convergence}
In a symmetric Markov game, suppose every step size $\alpha_t(s, \bar a)$ lies in $[0, 1]$ and, for every $s \in S$ and $a \in A$, with probability $1$,
\begin{equation}
\label{eq:lr-cond}
  \sum_{t} \alpha_t(s, \bar a) = \infty, \qquad \sum_{t} \alpha_t(s, \bar a)^2 \leq C,
\end{equation}
for a deterministic constant $C$. Then every agent's table $Q^i_t$ converges with probability $1$ to $Q^\ast$. Hence all agents converge to the same table.
\end{theorem}

\begin{proof}
Fix agent $i$ and let $Q_t := Q^i_t$. We apply~\citet[Theorem~3]{tsitsiklis1994} with $x(t) := Q_t$ and history $\mathcal{F}_t$, the history up to step $t$: the agents' initial tables and their exploration draws up to step $t$, the states, joint actions and rewards before step $t$, and the state $s_t$ and joint action at step $t$. On a step with $\mu_t = 1$, define the noise at the updated entry $(s_t, \bar a)$ as
\[
  w_t(s_t, \bar a) := r^i_t + \gamma \max_{a' \in A} Q_t(s_{t+1}, \bar a') - (HQ_t)(s_t, \bar a),
\]
and set $w_t(s, \bar a) := 0$ at every other entry. When $\mu_t = 0$, set $w_t(s, \bar a) := 0$ at every entry. Then entries evolve as
\[
  \begin{aligned}
  Q_{t+1}(s, \bar a) = Q_t(s, \bar a) &+ \alpha_t(s, \bar a) \bigl[ (HQ_t)(s, \bar a) \\
  &- Q_t(s, \bar a) + w_t(s, \bar a) \bigr],
  \end{aligned}
\]
since $\alpha_t(s, \bar a) = 0$ wherever $w_t(s, \bar a)$ was set to $0$. This is the asynchronous iteration of~\citet{tsitsiklis1994}, and we check the assumptions of its Theorem~3 in turn.

\noindent\textbf{Assumption~1} requires outdated values to be eventually discarded, and holds since every update uses the current table.

\noindent\textbf{Assumption~2} requires the following for every entry $(s, \bar a)$ and every $t$. We state each part in our notation and then check it.
\begin{enumerate}
  \item[(a)] \emph{$Q_0$ is $\mathcal{F}_0$-measurable.} This holds since $\mathcal{F}_0$ contains the initial tables.
  \item[(b)] \emph{$w_t(s, \bar a)$ is $\mathcal{F}_{t+1}$-measurable.} The noise is a function of $Q_t$, $s_t$, the joint action at step $t$, $r^i_t$ and $s_{t+1}$, all of which are determined by $\mathcal{F}_{t+1}$.
  \item[(c)] \emph{$\alpha_t(s, \bar a)$ is $\mathcal{F}_t$-measurable}, as are the delays, which here are zero. The match bit, and with it every step size, is determined by $s_t$, the joint action at step $t$ and the earlier update counts, all of which are determined by $\mathcal{F}_t$.
  \item[(d)] \emph{$\mathbb{E}[w_t(s, \bar a) \mid \mathcal{F}_t] = 0$.} On $\{\mu_t = 1\}$ the joint action is $\bar a$, so by the assumptions on the game
  \[
    \mathbb{E}\bigl[r^i_t + \gamma \max_{a'} Q_t(s_{t+1}, \bar a') \mid \mathcal{F}_t\bigr] = (HQ_t)(s_t, \bar a).
  \]
  Elsewhere $w_t = 0$.
  \item[(e)] \emph{There exist deterministic constants $A$ and $B$ such that}
  \[
    \mathbb{E}[w_t(s, \bar a)^2 \mid \mathcal{F}_t] \leq A + B \max_{t' \leq t} \| Q_{t'} \|^2.
  \]
  Both the sampled target and $(HQ_t)(s_t, \bar a)$ are at most $R_{\max} + \gamma \| Q_t \|$ in absolute value, so $|w_t| \leq 2R_{\max} + 2\gamma \| Q_t \|$. As $(x + y)^2 \leq 2x^2 + 2y^2$,
  \[
    w_t^2 \leq 8 R_{\max}^2 + 8 \gamma^2 \| Q_t \|^2,
  \]
  so~(e) holds with $A = 8 R_{\max}^2$ and $B = 8 \gamma^2$.
\end{enumerate}

\noindent\textbf{Assumptions~3 and~5.} Assumption~3 is the step-size condition~\eqref{eq:lr-cond}, and Assumption~5, that $H$ contracts towards its fixed point in a weighted max norm, holds with the sup norm by Lemma~\ref{lem:contraction}.

Theorem~3 of~\citet{tsitsiklis1994} therefore gives $Q^i_t \to Q^\ast$ with probability $1$. Since $H$, and with it $Q^\ast$, is the same for every agent, all agents converge to the same table.
\end{proof}

\noindent As~\citet{tsitsiklis1994} notes, Assumption~2(c) assumes that the samples can be generated after deciding which entries to update, which is what our algorithm takes advantage of.

\paragraph{The fixed point is the optimal Q-function of $M_K$.}
Theorem~\ref{thm:convergence} concerns training, during which the agents explore and play, and $M_K$ enters only through matched steps. If the optimal action is unique at every state, the players eventually act with the same greedy deterministic policy (Corollary~\ref{cor:play}). More generally, suppose every player follows the same deterministic policy $\pi$. All players observe the same state $s_t$, so they all play $\pi(s_t)$, and play stays on diagonal profiles. Every player's expected reward is then $r_K(s_t, \pi(s_t))$ and the state evolves by $P_K(\cdot \mid s_t, \pi(s_t))$, so the players are playing in $M_K$ itself. Since $H$ is the Bellman optimality operator of $M_K$, its fixed point $Q^\ast$ is the optimal Q-function of $M_K$~\citep{puterman1994markov}.

\begin{theorem}[Kant-Q learns the simple Kantian equilibrium]
\label{thm:kantq}
In a symmetric Markov game, a deterministic policy is a pure simple Kantian equilibrium if and only if it is optimal in $M_K$. In particular, a pure SKE exists. Let
\begin{equation}
\label{eq:greedy}
  \pi^\ast(s) := \arg\max_{a \in A} Q^\ast(s, \bar a), \qquad
  V^\ast(s) := \max_{a \in A} Q^\ast(s, \bar a),
\end{equation}
with ties broken by the lowest index. Then the greedy policy $\pi^\ast$ is a pure SKE.
\end{theorem}

\begin{proof}
By Definition~\ref{def:markov-ske}, a pure SKE is a best common policy in a Markov game, i.e., no player prefers any other common deterministic policy, from any state. By the construction of $M_K$ above, comparing common deterministic policies in the original game is the same as comparing deterministic policies of $M_K$, i.e., $V^i_{\pi}(s) = V^{M_K}_{\pi}(s)$ for every player $i$, state $s$ and deterministic policy $\pi$. A policy that is optimal in $M_K$ is thus a best common policy, and hence a pure SKE. Conversely, a pure SKE does at least as well as every deterministic policy of $M_K$, and $M_K$ has an optimal deterministic policy~\citep{puterman1994markov}, so a pure SKE attains the optimal value and is optimal in $M_K$. Finally, $\pi^\ast$ is greedy with respect to $Q^\ast$, so it is optimal in $M_K$, and in particular a pure SKE exists.
\end{proof}

Policies under which players randomise independently leave the diagonal and lie outside the scope of Definition~\ref{def:markov-ske}.

\begin{corollary}[Convergence of play]
\label{cor:play}
Suppose, in addition, that $a \mapsto Q^\ast(s, \bar a)$ has a unique maximiser at every state $s$. Then, with probability $1$, there is a finite time after which every agent's greedy action at every state $s$ is $\pi^\ast(s)$.
\end{corollary}

\noindent The proof is in Appendix~\ref{app:proofs}. The uniqueness assumption holds automatically in any single-state game whose diagonal payoffs $r(\bar a)$ are pairwise distinct, including every strictly monotone game. There $Q^\ast(\bar a) = r(\bar a) + \gamma V^\ast$. Without uniqueness, an exact tie in $Q^\ast$ can leave the agents' estimates swapping order forever and differing across agents. The tables still converge, but greedy play can miscoordinate indefinitely.

\subsection{Shared exploration}
\label{sec:shared}

Under independent $\varepsilon$-greedy exploration, an entry that no agent is greedy on is updated with probability $(\varepsilon / m)^N$ per visit. \emph{Shared exploration} removes this factor. Before play, the agents agree on a common random seed. Each agent seeds its own copy of a random number generator with it and, at every step, draws its action uniformly at random from $A$ using that generator, that is, it runs Algorithm~\ref{alg:kantq} with $\varepsilon = 1$. The tables are still initialised independently. This way, all agents play the same action at every step, so $\mu_t = 1$ at every step and every step is an update. No information is exchanged during play. Each visit to a state now updates each of its entries with probability $1/m$ instead of $(\varepsilon / m)^N$.

Theorem~\ref{thm:convergence} applies unchanged, since each visit to a state plays a uniformly random common action, so given every state is visited infinitely often, every $(s, \bar a)$ is updated infinitely often. Since play never leaves the diagonal, states that cannot be reached by diagonal play are visited only if play restarts there, so we let play restart infinitely often from a state drawn from a distribution with full support on $S$.

\paragraph{Convergence rate.}
Under shared exploration every step is a matched step, so each agent's experience is a sequence of samples from $M_K$ with uniformly random actions, and its updates are exactly the asynchronous Q-learning of~\citet{even-dar}. Their convergence rates therefore apply. Following them, the \emph{covering time} $L$ is a number of steps such that, from any start state, with probability at least $\tfrac12$ every pair $(s, a)$ is played within $L$ steps. Let $\beta := (1 - \gamma)/2$.

\begin{proposition}[Convergence rate with shared exploration]
\label{prop:rate}
In a symmetric Markov game with realised rewards in $[0, R_{\max}]$, let $V_{\max} := R_{\max} / (1 - \gamma)$, and suppose every initial table entry lies in $[0, V_{\max}]$. Let the agents use shared exploration with the rate $n^{-\omega}$ of Algorithm~\ref{alg:kantq}. Then there is a constant $c > 0$ such that, for every $\eta, \delta \in (0, 1)$ and every agent $i$, with probability at least $1 - \delta$, $\| Q^i_T - Q^\ast \| \leq \eta$, provided $T \geq c\,(T_1 + T_2)$, where
\[
  \begin{aligned}
  T_1 &:= \Bigl( \frac{\tilde L^{1+3\omega} V_{\max}^2}{\beta^2 \eta^2} \ln \frac{|S| \, m \, V_{\max}}{\delta \beta \eta} \Bigr)^{1/\omega}, \\
  T_2 &:= \Bigl( \frac{\tilde L}{\beta} \ln \frac{V_{\max}}{\eta} \Bigr)^{1/(1-\omega)},
  \end{aligned}
\]
and $\tilde L := L \lceil \log_2(T / \delta) \rceil$.
\end{proposition}

\begin{proof}
Every step is a matched step, so agent $i$'s experience tuples are a sequence of samples from $M_K$ with uniformly random actions, with rewards in $[0, R_{\max}]$. Its rate at entry $(s, \bar a)$ is $n^{-\omega}$, where $n$ is the number of updates of that entry so far, including this one, so agent $i$ runs the asynchronous Q-learning of~\citet[Section~3]{even-dar} on $M_K$. Their analysis does not assume that the samples form a single trajectory, so restarts are allowed. By~\citet[Corollary~34]{even-dar}, with probability at least $1 - \delta$, every pair is played within every window of $\tilde L$ steps of a run of length $T$, so the run has covering time $\tilde L$. On this event, \citet[Theorem~4]{even-dar} gives $\| Q^i_T - Q^\ast \| \leq \eta$ with probability at least $1 - \delta$. Applying a union bound over these two events and rescaling $\delta$ gives the stated result.
\end{proof}

\noindent The condition on $T$ is a lower bound in which $T$ also appears on the right, through $\tilde L$. Since $\tilde L$ grows only logarithmically in $T$, $T_1 + T_2$ grows only polylogarithmically in $T$, so such a $T$ always exists, and the condition holds for all sufficiently large $T$.

\paragraph{Training and deployment.}
Under this shared exploration, play does not converge as every action is uniformly random. In practice, the agents explore for a fixed budget of $T$ steps and then all act greedily. By Corollary~\ref{cor:play}, with probability $1$ there is a finite time after which every agent's greedy action  is $\pi^\ast(s)$ at every state. If $T$ is past that time, deployed play follows $\pi^\ast$ and stays on the diagonal. Proposition~\ref{prop:rate} defines the budget. Under the uniqueness assumption of Corollary~\ref{cor:play}, let $\Delta > 0$ be the smallest gap between $Q^\ast(s, \overline{\pi^\ast(s)})$ and $Q^\ast(s, \bar a)$, over states $s$ and actions $a \neq \pi^\ast(s)$. If $T$ satisfies the bound of Proposition~\ref{prop:rate} with $\eta < \Delta / 2$ and with $\delta / N$ in place of $\delta$, then, by the argument of Corollary~\ref{cor:play}, with probability at least $1 - \delta$ every agent's greedy action after $T$ steps is $\pi^\ast(s)$ at every state $s$. Using $\delta / N$ makes the bound hold for all agents at once, at the cost of polylogarithmic terms in $N$. Apart from this, for a fixed $M_K$ the bound does not depend on $N$ at all.

\section{Experiments}
\label{sec:experiments}

We test the two claims of the previous section, that every agent's table converges to $Q^\ast$, and that the agents' greedy policies are then the simple Kantian equilibrium. We compare Kant-Q with independent Q-learning (IQL), team IQL (trained on the average team reward), Friend-Q and value iteration on $M_K$. All learners initialise their tables independently and uniformly at random in $[0, 1]$ and use the rate $n^{-\omega}$ with $\omega = 0.7$. Each experiment is run with 5 seeds, and all metrics are averaged over them. In the tables, \emph{learnt SKE} is ``yes'' when every agent's final greedy policy is $\pi^\ast$ in every seed. Value iteration solves $M_K$ exactly, so it gives $Q^\ast$ and the simple Kantian equilibrium $\pi^\ast$ (Theorem~\ref{thm:kantq}), the optimal target for Kant-Q.

\subsection{Iterated Prisoner's Dilemma}

\begin{table}[h]
  \centering
  \begin{tabular}{lcc}
    \toprule
     & $C$ & $D$ \\
    \midrule
    $C$ & $3, 3$ & $0, 5$ \\
    $D$ & $5, 0$ & $1, 1$ \\
    \bottomrule
  \end{tabular}
  \caption{Payoffs of the Prisoner's Dilemma stage game (row player, column player).}
  \label{tab:pd}
\end{table}

Table~\ref{tab:pd} gives the stage game. The state is the previous round's joint action, play starts at $CC$, and $\gamma = 0.9$. As a Markov game, it has $N = 2$ players, actions $A = \{C, D\}$ and states $S = \{CC, CD, DC, DD\}$. Playing the joint action $\mathbf{b}$ moves the game deterministically to state $\mathbf{b}$, and the rewards at every state are the stage payoffs of Table~\ref{tab:pd}, so the game is symmetric. The SKE $\pi^\ast$ is to always cooperate. The agents explore independently with $\varepsilon = 0.5$ for $3 \cdot 10^5$ steps.

\begin{figure}[t]
  \centering
  \includegraphics[width=\columnwidth]{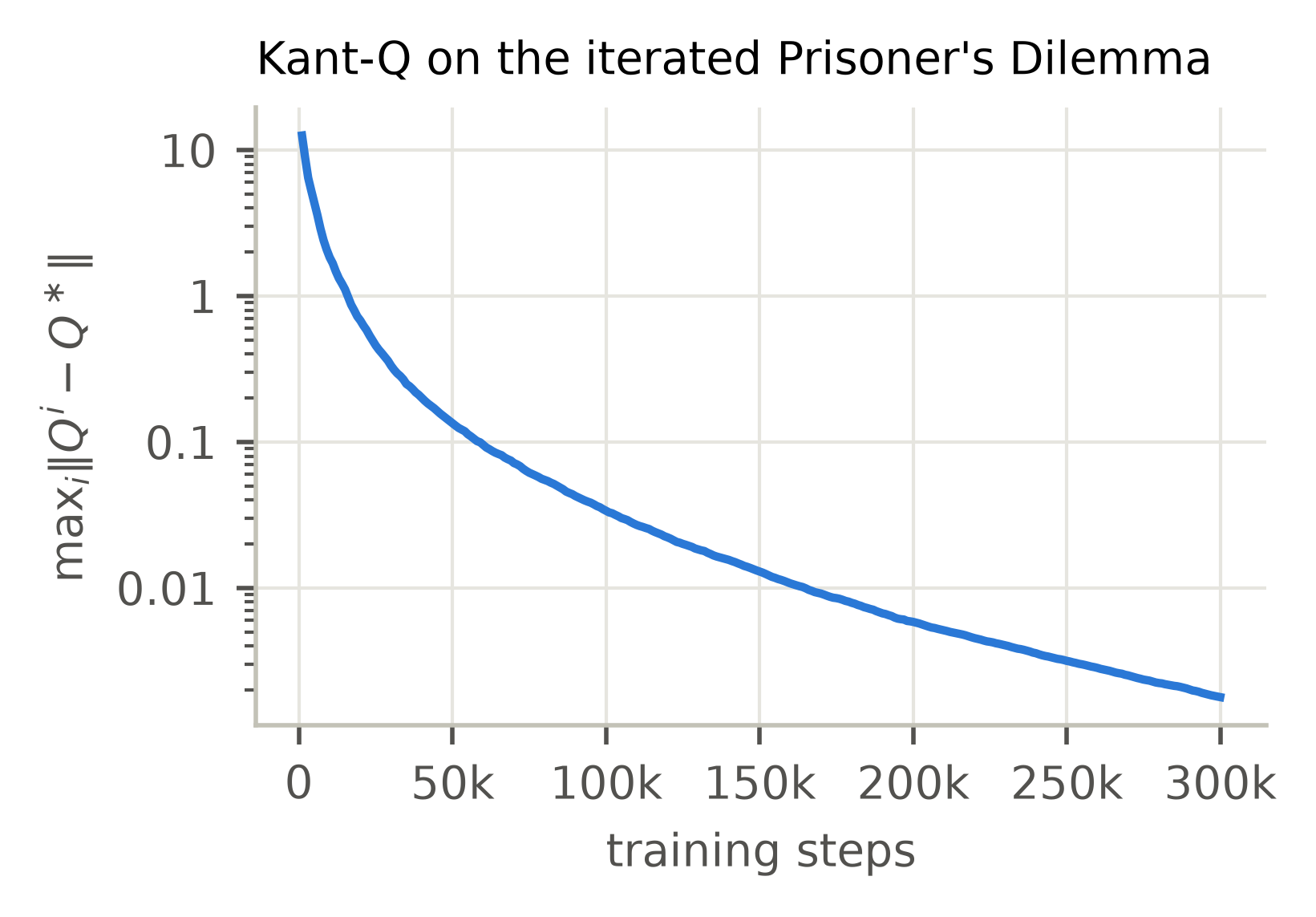}
  \caption{Convergence of Kant-Q on the iterated Prisoner's Dilemma.}
  \label{fig:ipd}
\end{figure}

The error of Kant-Q falls steadily to below $2 \cdot 10^{-3}$ in every seed (Figure~\ref{fig:ipd}), as Theorem~\ref{thm:convergence} predicts, and every agent's greedy policy is $\pi^\ast$ (Table~\ref{tab:ipd}). IQL and Friend-Q learn to always defect and earn the Nash payoff $1$ per round. Team IQL also cooperates, since the mean reward makes cooperation dominant in the Prisoner's Dilemma.

\begin{table}[h]
  \centering
  \begin{tabular}{lcc}
\toprule
 & learned policy & learnt SKE \\
\midrule
Value iteration on $M_K$ & Always-C & --- \\
Kant-Q & Always-C & yes \\
IQL & Always-D & no \\
Team IQL & Always-C & yes \\
Friend-Q & Always-D & no \\
\bottomrule
\end{tabular}

  \caption{Learned strategies on the iterated Prisoner's Dilemma.}
  \label{tab:ipd}
\end{table}

\begin{table*}[t]
  \centering
  \begin{tabular}{llcccccc}
\toprule
N & algorithm & learnt SKE & catch at full lake & months survived & catch / agent & equality & over-usage \\
\midrule
5 & Value iteration on $M_K$ & --- & 10.0 & 12.0 & 120 & 1.00 & 0\% \\
5 & Kant-Q (shared) & yes & 10.0 & 12.0 & 120 & 1.00 & 0\% \\
5 & Kant-Q (independent) & no & 26.2 & 5.4 & 50 & 0.95 & 62\% \\
5 & IQL & no & 48.4 & 1.0 & 20 & 0.90 & 99\% \\
5 & Team IQL & no & 0.2 & 12.0 & 2 & 0.84 & 0\% \\
5 & Team IQL ($\varepsilon = 0.1$) & no & 9.4 & 12.0 & 113 & 0.70 & 16\% \\
5 & Friend-Q & no & 42.4 & 1.0 & 20 & 0.90 & 99\% \\
\midrule
200 & Value iteration on $M_K$ & --- & 10.0 & 12.0 & 120 & 1.00 & 0\% \\
200 & Kant-Q (shared) & yes & 10.0 & 12.0 & 120 & 1.00 & 0\% \\
200 & Kant-Q (independent) & no & 22.1 & 1.0 & 20 & 0.61 & 64\% \\
200 & IQL & no & 45.5 & 1.0 & 20 & 0.87 & 99\% \\
200 & Team IQL & no & 18.8 & 5.4 & 48 & 0.59 & 52\% \\
200 & Team IQL ($\varepsilon = 0.1$) & no & 28.1 & 1.0 & 20 & 0.74 & 79\% \\
\bottomrule
\end{tabular}

  \caption{Learned strategies and outcomes on the GovSim fishery.}
  \label{tab:govsim}
\end{table*}

\subsection{GovSim fishery}
\label{sec:govsim}

GovSim was introduced to test whether LLM agents can sustain a common-pool resource~\citep{govsim}. We use a tabular reimplementation of their fishery. Each month, every agent asks for a catch from $\{0, 5, 10, 15, 20, 30, 40, 50\}$ tons. If the lake holds enough fish, every agent gets what it asked for. Otherwise the fish are handed out one ton at a time to a uniformly random agent which is still requesting more, which empties the lake, as in GovSim. If fewer than $N$ tons remain, the lake collapses, an absorbing state with zero reward. Otherwise the stock, counted in whole tons per agent, doubles, up to the capacity $20N$. For $N = 5$ these are exactly GovSim's settings. The state is the stock per agent before the harvest, which all agents share. It takes the values $2, 4, \dots, 20$ tons per agent or the collapsed lake, giving $11$ states across values of $N$. The actions are the $8$ catch sizes, and each agent's reward is the catch it receives. The next state is determined by the current stock and the total catch requested, so the only randomness is in the rewards, when the lake cannot meet every request. Since the fish are then handed out uniformly at random, the expected rewards are symmetric, and the fishery is a symmetric Markov game. GovSim's 12-month limit is a reset, and the agents learn the infinite-horizon discounted value with $\gamma = 0.9$. Episodes start from a uniformly random stock, so that shared exploration, which never leaves the diagonal, visits every state.

Kant-Q is trained with shared exploration (Section~\ref{sec:shared}) and with independent $\varepsilon$-greedy exploration, $\varepsilon = 0.5$ as in the IPD, the baselines with the latter, each for $10^6$ steps. A smaller $\varepsilon$ helps team IQL for $N = 5$, so Table~\ref{tab:govsim} also gives it with $\varepsilon = 0.1$. Friend-Q's table has $8^N$ entries per state, so it is run only for $N = 5$. After training, the agents play GovSim's 12-month episode from a full lake, averaged over 200 rollouts per seed, and we report the following GovSim metrics.
\begin{itemize}
    \item \textbf{Survival time.} The number of months harvested, counting the month in which the lake collapses ($12$ if it never does).
    \item \textbf{Catch per agent.} Each agent's total catch across 12 months.
    \item \textbf{Equality.} One minus the Gini coefficient of the agents' catches.
    \item \textbf{Over-usage.} The fraction of agent-months in which an agent catches more than the sustainability threshold $\lfloor \lfloor \text{stock}/2 \rfloor / N \rfloor$.
\end{itemize}

\begin{figure}[t]
  \centering
  \includegraphics[width=\columnwidth]{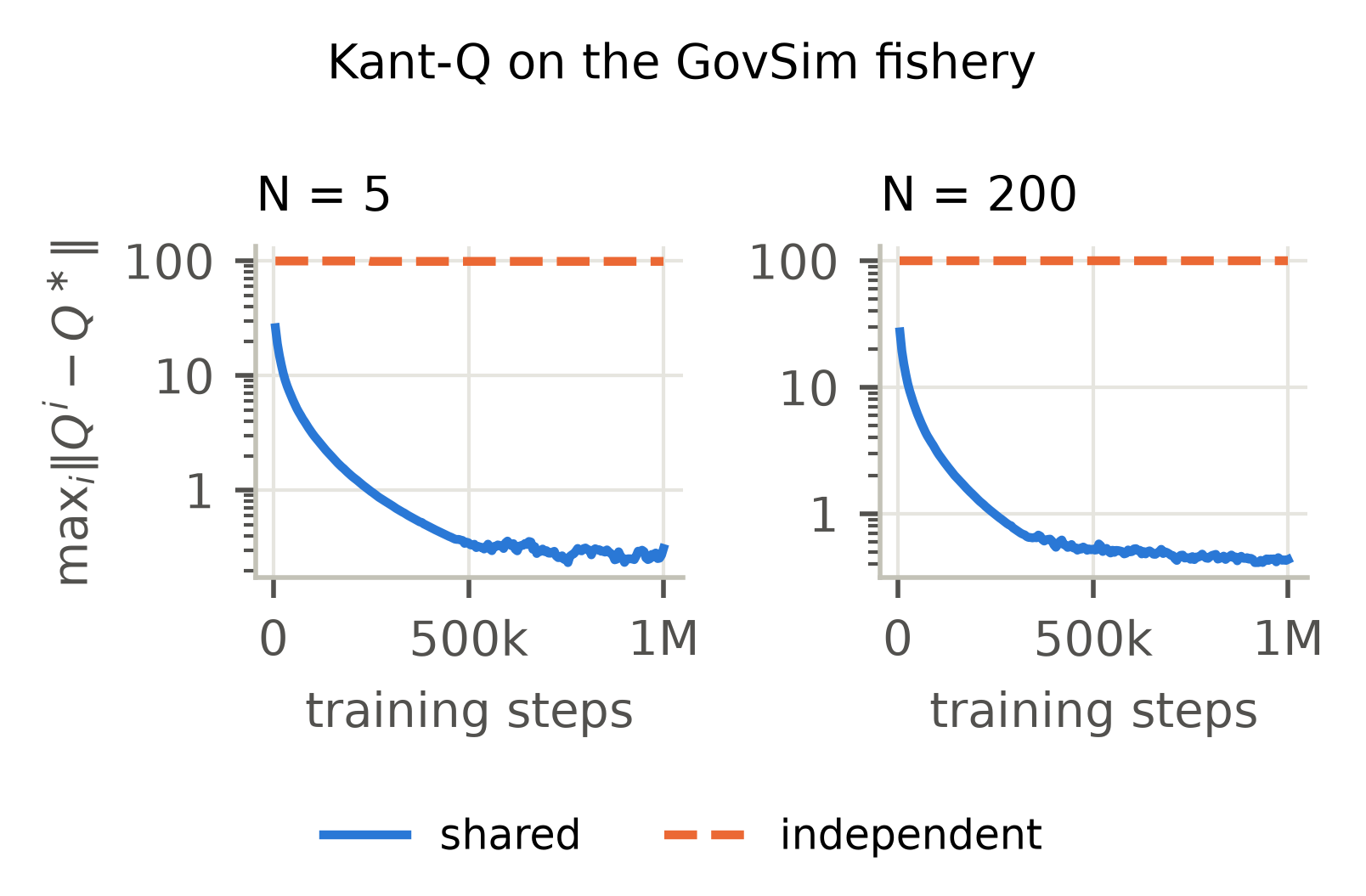}
  \caption{Convergence of Kant-Q on the GovSim fishery.}
  \label{fig:govsim}
\end{figure}

With shared exploration, the error of Kant-Q falls at every $N$, while with independent exploration it stays near its initial value (Figure~\ref{fig:govsim}). Independent tables almost never agree on an action, so matched steps and updates are rare. Since $M_K$ is the same for every $N$, the error falls at similar rates for $N = 5$ and $N = 200$, as Proposition~\ref{prop:rate} predicts. With shared exploration, deployed agents match value iteration exactly. The lake survives all 12 months, every agent catches $120$ tons, and no agent over-uses it. IQL and Friend-Q ask for $42$--$48$ tons on average from a full lake, and the lake collapses in the first month. Team IQL's performance depends heavily on $\varepsilon$. With $\varepsilon = 0.1$, team IQL comes close to the simple Kantian equilibrium for $N = 5$. The lake survives all 12 months and the agents catch $113$ tons on average. But the catch is not shared equally (equality $0.70$), and agents over-use the lake in $16\%$ of agent-months. The team reward depends only on the total catch, so team IQL learns how much the group should take but the agents settle on an unequal split. Team IQL thus optimises a utilitarian objective. It values the total catch but is indifferent to how the catch is divided. Kant-Q learns one action for all agents, so every agent catches the same $120$ tons. Appendix~\ref{app:govsim} gives the results for $N = 10$ and $N = 100$, and for every baseline at $\varepsilon = 0.1$, $0.3$ and $0.5$.

\section{Discussion and Limitations}
\label{sec:limitations}

\paragraph{Kantian equilibrium as an economic solution concept.}
As discussed in the introduction, the Kantian equilibrium is an economic solution concept that offers a fair alternative to welfare maximisation. The GovSim experiments show the contrast with a utilitarian objective. With $\varepsilon = 0.1$, team IQL, which maximises the total reward, learns roughly how much the group should catch but splits it unequally (equality $0.70$ for $N = 5$), while Kant-Q gives every agent the same catch. Comparing the Kantian equilibrium with other welfare criteria, such as utilitarian, egalitarian and Nash welfare, is a natural direction for future work.

\paragraph{Designing AI--AI interaction.}
Kant-Q is a natural design when the interacting agents are copies of one AI system, for example a fleet of agents drawing on a shared resource. No centralised training is needed. Each agent learns from its own reward and the match bit alone, unlike methods trained centrally on joint information, such as Multi-Agent PPO~\citep{MAPPO}, or on a shared team reward. The same holds when the agents represent large numbers of people, for example people drawing on the fishery of Section~\ref{sec:govsim}, where Kant-Q acts as a mechanism that reaches a sustainable, equal outcome without a central planner. This requires the people's preferences to be symmetric and every representative to run Kant-Q, which the next two paragraphs discuss.

\paragraph{Symmetric games and shared situations.}
Kant-Q asks which action every agent would want all agents to take. This presupposes a common action set, agents that value the diagonal profiles identically, and a state that all agents share, so that a common policy prescribes the same action to every agent. In asymmetric games, ``everyone plays the same action'' is no longer a meaningful counterfactual. \citet{roemer} defines additive and multiplicative Kantian equilibria for heterogeneous agents, in which every agent changes its strategy by the same factor or the same amount. These are defined for continuous strategy spaces, and extending Kant-Q to them is a natural next step. When agents instead observe the state from their own perspective, as in the usual formulation of the iterated Prisoner's Dilemma, a common policy such as tit-for-tat can prescribe different actions to different agents, and play leaves the diagonal. The Kantian question then becomes which \emph{policy} every agent would want all agents to follow, which Kant-Q's per-action table does not represent. Universalising policies rather than actions is the main open direction.

\paragraph{Vulnerability to exploitation.}
This matters for AI--AI interaction, where agents from different developers need not share an algorithm. Kant-Q assumes that the other agents reason the same way. An agent that does not, for example one that always defects, breaks this assumption. A Kant-Q agent learns only from steps on which every agent matched its action, so it receives no updates for the actions the defector never plays, and its greedy action is determined by its initial values for those entries. It does not detect or punish the defector, unlike methods such as FCL~\citep{fcl} that retaliate against deviations. Combining Kant-Q with a test for whether the other agents are Kantian, and a fallback to a best response when they are not, would make it robust to such agents.

\bibliography{refs}

\appendix

\section{Proofs}
\label{app:proofs}

\subsection{Individual rationality of the SKE}

The proof of Proposition~\ref{prop:ir} uses the following lemma.

\begin{lemma}[Extremal Opponent Profile]
\label{lem:extremal}
In a strictly monotone decreasing game with action set $A = \{a_1, \ldots, a_m\}$, for every $b \in A$ and every opponent profile $c \in A^{N-1}$ with $c \neq (a_m, \ldots, a_m)$,
\[
    r(b \, ; c) > r(b \, ; a_m, \ldots, a_m).
\]
\end{lemma}

\begin{proof}
Let $c = (c_1, \ldots, c_{N-1})$. We switch the opponents to $a_m$ one at a time. For $k = 0, 1, \ldots, N-1$, let
\[
    c^{(k)} := (\underbrace{a_m, \ldots, a_m}_{k}, c_{k+1}, \ldots, c_{N-1}),
\]
so that $c^{(0)} = c$ and $c^{(N-1)} = (a_m, \ldots, a_m)$.

Consecutive profiles $c^{(k-1)}$ and $c^{(k)}$ differ only in coordinate $k$, which changes from $c_k$ to $a_m$. If $c_k = a_m$ the two profiles coincide. Otherwise $c_k = a_j$ for some $j < m$, and the Monotonicity clause of Definition~\ref{def:monotone} gives
\[
    r(b \, ; c^{(k-1)}) > r(b \, ; c^{(k)}).
\]
Hence $r(b \, ; c^{(k-1)}) \geq r(b \, ; c^{(k)})$ for every $k$, and the inequality is strict for at least one $k$, since $c \neq (a_m, \ldots, a_m)$. Chaining these $N-1$ inequalities gives
\[
    \begin{aligned}
    r(b \, ; c) = r(b \, ; c^{(0)}) &> r(b \, ; c^{(N-1)}) \\
    &= r(b \, ; a_m, \ldots, a_m).
    \end{aligned}
\]
\end{proof}

\begin{proof}[Proof of Proposition~\ref{prop:ir}]
We prove the result for the strictly monotone decreasing case. The increasing case follows by relabelling the action indices. The unique SKE is then $a_1$. Let $p^* = r(\bar a_1)$ denote each player's payoff when all play it.

Following~\citet{myerson}, the minmax value of player $i$ in pure strategies is
\[
    v^i = \min_{c \, \in \, A^{N-1}} \; \max_{b \, \in \, A} \; r(b \, ; c).
\]
By symmetry this quantity is the same for every player, so it suffices to bound it for player $1$, and we let $v := v^1$.

\medskip
\noindent\textbf{Step 1. The minmax is attained when all opponents play $a_m$.}
By Lemma~\ref{lem:extremal}, $r(b \, ; c) > r(b \, ; a_m, \ldots, a_m)$ for every $b \in A$ and every $c \neq (a_m, \ldots, a_m)$. Taking the maximum over $b$ on both sides,
\[
    \max_{b \in A} r(b \, ; c) \; > \; \max_{b \in A} r(b \, ; a_m, \ldots, a_m)
\]
for every such $c$. Therefore the outer minimum is attained uniquely at $c = (a_m, \ldots, a_m)$, and
\[
    v = \max_{b \, \in \, A} \; r(b \, ; a_m, \ldots, a_m).
\]

\medskip
\noindent\textbf{Step 2. $p^*$ strictly exceeds that maximum.}
If $b \neq a_m$, Lemma~\ref{lem:extremal} with $c = (b, \ldots, b)$ and the Common Diagonal Ordering clause give
\[
    r(b \, ; a_m, \ldots, a_m) < r(b \, ; b, \ldots, b) = r(\bar b) \leq r(\bar a_1) = p^*.
\]
If $b = a_m$, then $r(b \, ; a_m, \ldots, a_m) = r(\bar a_m) < r(\bar a_1) = p^*$ by the Common Diagonal Ordering clause, since $m \geq 2$. Hence $r(b \, ; a_m, \ldots, a_m) < p^*$ for every $b$, and taking the maximum over $b$ gives $v < p^*$.
\end{proof}

\noindent We use the minmax value since it is the standard threshold for individual rationality. The pure maxmin (security) value would make the result nearly trivial. In any symmetric game with an SKE $a^*$, for every $b \in A$,
\[
    \min_{c \, \in \, A^{N-1}} r(b \, ; c) \;\leq\; r(\bar b) \;\leq\; r(\bar a^*),
\]
so taking the maximum over $b$, the SKE payoff weakly exceeds the maxmin value.

\subsection{Contraction of the Kant-Q operator}

\begin{proof}[Proof of Lemma~\ref{lem:contraction}]
Fix $s \in S$ and $a \in A$. By~\eqref{eq:kantq-op},
\[
  \begin{aligned}
  |(HQ)(s, \bar a) - (HQ')(s, \bar a)| &\leq \gamma \sum_{s' \in S} P(s' \mid s, \bar a) \, \Delta(s') \\
  &\leq \gamma \| Q - Q' \|,
  \end{aligned}
\]
where $\Delta(s') := |\max_{a'} Q(s', \bar a') - \max_{a'} Q'(s', \bar a')| \leq \| Q - Q' \|$, since the difference of two maxima is at most the largest difference, and $P(\cdot \mid s, \bar a)$ sums to $1$. This holds for every $s$ and $a$, so $\| HQ - HQ' \| = \max_{s, a} |(HQ)(s, \bar a) - (HQ')(s, \bar a)| \leq \gamma \| Q - Q' \|$. Since $\mathcal{Q}$ with the sup norm is complete, the Banach fixed-point theorem~\citep{Banach} gives a unique fixed point.
\end{proof}

\subsection{Convergence of play}

\begin{proof}[Proof of Corollary~\ref{cor:play}]
Write $b_s := \pi^\ast(s)$, and let $\Delta > 0$ be the smallest gap, over states $s$ and actions $a \neq b_s$, between $Q^\ast(s, \bar b_s)$ and $Q^\ast(s, \bar a)$. By Theorem~\ref{thm:convergence}, with probability $1$ there is a finite $T$ such that $\| Q^i_t - Q^\ast \| < \Delta / 2$ for every agent $i$ and every $t \geq T$. For such $t$, every state $s$ and every $a \neq b_s$,
\[
  Q^i_t(s, \bar b_s) > Q^\ast(s, \bar b_s) - \tfrac{\Delta}{2} \geq Q^\ast(s, \bar a) + \tfrac{\Delta}{2} > Q^i_t(s, \bar a),
\]
so the greedy action at $s$ is $b_s$.
\end{proof}

\section{GovSim results for every $N$}
\label{app:govsim}

\begin{figure}[t]
  \centering
  \includegraphics[width=\columnwidth]{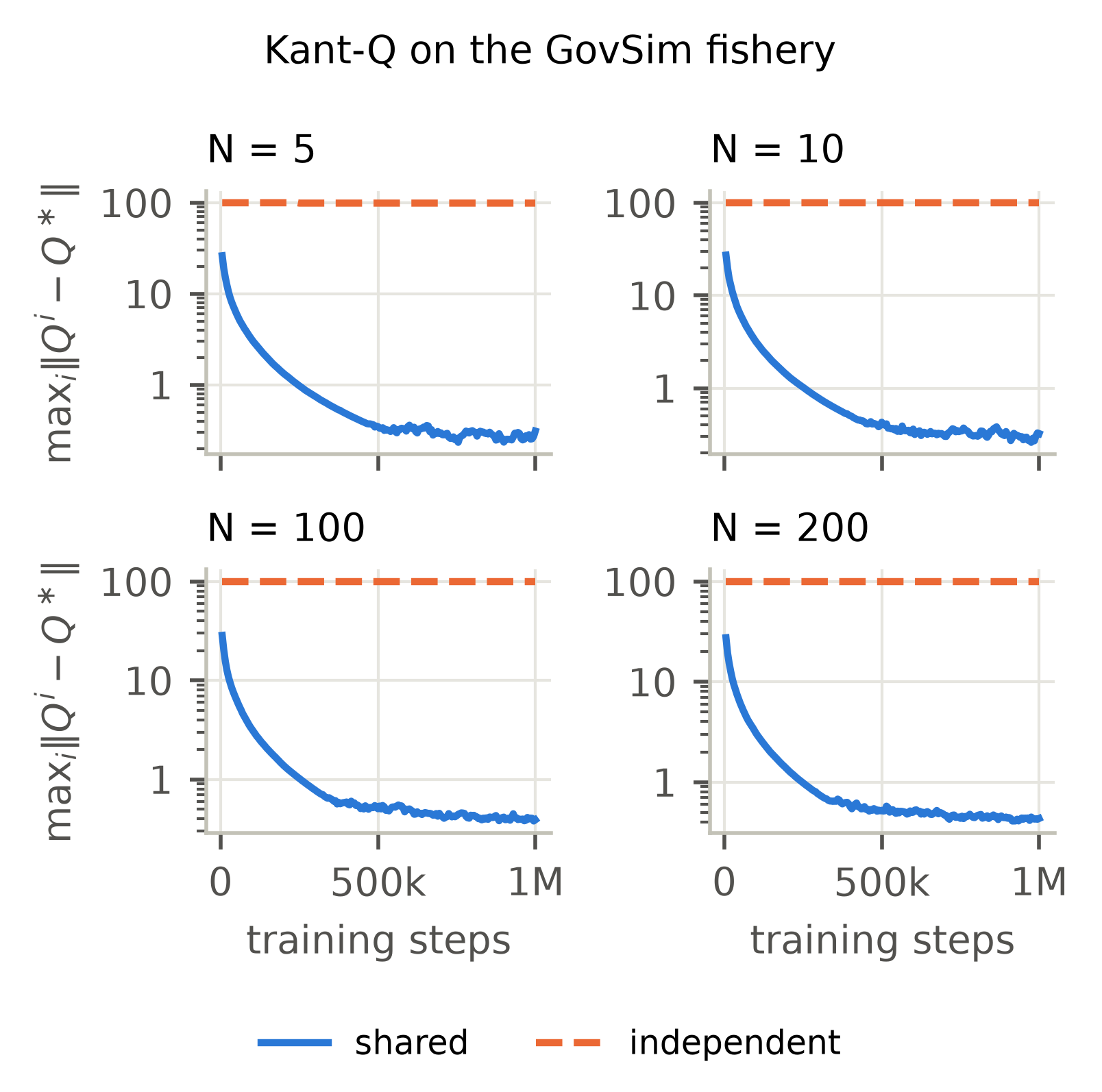}
  \caption{Convergence of Kant-Q on the GovSim fishery for every $N$.}
  \label{fig:govsim-all}
\end{figure}

Figure~\ref{fig:govsim-all} and Table~\ref{tab:govsim-all} extend Figure~\ref{fig:govsim} and Table~\ref{tab:govsim} to $N = 10$ and $N = 100$. Table~\ref{tab:govsim-all} also gives each baseline at $\varepsilon = 0.1$ and $0.3$, besides the $\varepsilon = 0.5$ of the main runs. Kant-Q with shared exploration uses $\varepsilon = 1$.

\begin{table*}[t]
  \centering
  \setlength{\tabcolsep}{4pt}
  \begin{tabular}{lllcccccc}
\toprule
N & algorithm & $\varepsilon$ & learnt SKE & catch at full lake & months survived & catch / agent & equality & over-usage \\
\midrule
5 & Value iteration on $M_K$ & --- & --- & 10.0 & 12.0 & 120 & 1.00 & 0\% \\
5 & Kant-Q (independent) & 0.5 & no & 26.2 & 5.4 & 50 & 0.95 & 62\% \\
5 & Kant-Q (shared) & 1.0 & yes & 10.0 & 12.0 & 120 & 1.00 & 0\% \\
5 & IQL & 0.1 & no & 42.8 & 1.0 & 20 & 0.90 & 99\% \\
5 & IQL & 0.3 & no & 45.6 & 1.0 & 20 & 0.90 & 99\% \\
5 & IQL & 0.5 & no & 48.4 & 1.0 & 20 & 0.90 & 99\% \\
5 & Team IQL & 0.1 & no & 9.4 & 12.0 & 113 & 0.70 & 16\% \\
5 & Team IQL & 0.3 & no & 5.2 & 12.0 & 62 & 0.63 & 4\% \\
5 & Team IQL & 0.5 & no & 0.2 & 12.0 & 2 & 0.84 & 0\% \\
5 & Friend-Q & 0.1 & no & 27.8 & 1.2 & 21 & 0.87 & 93\% \\
5 & Friend-Q & 0.3 & no & 38.4 & 1.0 & 20 & 0.90 & 99\% \\
5 & Friend-Q & 0.5 & no & 42.4 & 1.0 & 20 & 0.90 & 99\% \\
\midrule
10 & Value iteration on $M_K$ & --- & --- & 10.0 & 12.0 & 120 & 1.00 & 0\% \\
10 & Kant-Q (independent) & 0.5 & no & 21.5 & 1.2 & 20 & 0.67 & 69\% \\
10 & Kant-Q (shared) & 1.0 & yes & 10.0 & 12.0 & 120 & 1.00 & 0\% \\
10 & IQL & 0.1 & no & 41.6 & 1.0 & 20 & 0.89 & 99\% \\
10 & IQL & 0.3 & no & 43.0 & 1.0 & 20 & 0.89 & 99\% \\
10 & IQL & 0.5 & no & 48.2 & 1.0 & 20 & 0.89 & 99\% \\
10 & Team IQL & 0.1 & 2\% of agents & 8.9 & 12.0 & 107 & 0.64 & 24\% \\
10 & Team IQL & 0.3 & no & 2.5 & 12.0 & 30 & 0.30 & 4\% \\
10 & Team IQL & 0.5 & no & 0.0 & 12.0 & 0 & 1.00 & 0\% \\
\midrule
100 & Value iteration on $M_K$ & --- & --- & 10.0 & 12.0 & 120 & 1.00 & 0\% \\
100 & Kant-Q (independent) & 0.5 & no & 22.3 & 1.0 & 20 & 0.61 & 63\% \\
100 & Kant-Q (shared) & 1.0 & yes & 10.0 & 12.0 & 120 & 1.00 & 0\% \\
100 & IQL & 0.1 & no & 41.1 & 1.0 & 20 & 0.88 & 99\% \\
100 & IQL & 0.3 & no & 44.4 & 1.0 & 20 & 0.88 & 99\% \\
100 & IQL & 0.5 & no & 45.5 & 1.0 & 20 & 0.88 & 99\% \\
100 & Team IQL & 0.1 & no & 23.6 & 3.2 & 32 & 0.67 & 66\% \\
100 & Team IQL & 0.3 & no & 5.6 & 10.0 & 40 & 0.42 & 18\% \\
100 & Team IQL & 0.5 & no & 1.0 & 12.0 & 12 & 0.14 & 1\% \\
\midrule
200 & Value iteration on $M_K$ & --- & --- & 10.0 & 12.0 & 120 & 1.00 & 0\% \\
200 & Kant-Q (independent) & 0.5 & no & 22.1 & 1.0 & 20 & 0.61 & 64\% \\
200 & Kant-Q (shared) & 1.0 & yes & 10.0 & 12.0 & 120 & 1.00 & 0\% \\
200 & IQL & 0.1 & no & 41.4 & 1.0 & 20 & 0.88 & 99\% \\
200 & IQL & 0.3 & no & 45.0 & 1.0 & 20 & 0.87 & 99\% \\
200 & IQL & 0.5 & no & 45.5 & 1.0 & 20 & 0.87 & 99\% \\
200 & Team IQL & 0.1 & no & 28.1 & 1.0 & 20 & 0.74 & 79\% \\
200 & Team IQL & 0.3 & no & 21.4 & 3.2 & 24 & 0.62 & 59\% \\
200 & Team IQL & 0.5 & no & 18.8 & 5.4 & 48 & 0.59 & 52\% \\
\bottomrule
\end{tabular}

  \caption{Learned strategies and outcomes on the GovSim fishery for every $N$, with each baseline at every exploration rate $\varepsilon$.}
  \label{tab:govsim-all}
\end{table*}

\end{document}